\documentclass[sigconf]{acmart}
\usepackage{tikz}
\usepackage{pifont}
\usepackage{wasysym}
\usepackage{mathtools}
\usepackage{graphicx}
\usepackage{caption, subcaption}
\usepackage{url}
\usepackage{amsmath, amsfonts, amsthm, bm}
\usepackage{tabularx, multirow}
\usepackage[ruled,vlined]{algorithm2e}
\usepackage{booktabs}
\usepackage{xcolor, colortbl}
\usepackage{xspace}
\usepackage{resizegather}
\usepackage{makecell}
\usepackage{enumitem}

\usepackage{ltxtable}
\usepackage{cuted} 
\usepackage{supertabular}

\newcolumntype{P}{>{\raggedright\arraybackslash}m{0.95\linewidth}}
\newcolumntype{Q}{>{\centering\arraybackslash}m{0.11\linewidth}}
\newcolumntype{R}{>{\raggedright\arraybackslash}m{0.12\linewidth}}
\newcolumntype{S}{>{\centering\arraybackslash}m{0.06\linewidth}}
\newcolumntype{U}{>{\centering\arraybackslash}m{0.12\linewidth}}

\AtBeginDocument{%
  }

\copyrightyear{2026}
\acmYear{2026}
\setcopyright{cc}
\setcctype{by}
\acmConference[SIGIR '26]{Proceedings of the 49th International ACM SIGIR Conference on Research and Development in Information Retrieval}{July 20--24, 2026}{Melbourne, VIC, Australia}
\acmBooktitle{Proceedings of the 49th International ACM SIGIR Conference on Research and Development in Information Retrieval (SIGIR '26), July 20--24, 2026, Melbourne, VIC, Australia}
\acmDOI{10.1145/3805712.3809674}
\acmISBN{979-8-4007-2599-9/2026/07}

\begin{document}

\newcommand{\bench}{\textsc{IPQA}\xspace}
\newcommand{\eval}{\textsc{IPQA-Eval}\xspace}

\newcommand{\nopers}{No Personalization\xspace}
\newcommand{\ranicl}{Random Profile ICL\xspace}
\newcommand{\rawrag}{Profile RAG (Raw)\xspace}
\newcommand{\intrag}{Profile RAG (Intents)\xspace}

\newcommand{\pqa}{PQA\xspace}

\definecolor{DarkGreen}{RGB}{30,130,30}
\newcommand{\cmark}{\textcolor{DarkGreen}{\ding{51}}}
\newcommand{\xmark}{\textcolor{red}{\ding{55}}}%

% Camera-ready revision highlighter. Toggle off by redefining to: \newcommand{\rev}[1]{\textcolor{blue}{#1}}
\newcommand{\rev}[1]{#1}

% Unmarked footnote (for corresponding-author note placed via abstract)
\newcommand\blfootnote[1]{%
  \begingroup
  \renewcommand\thefootnote{}\footnote{#1}%
  \addtocounter{footnote}{-1}%
  \endgroup
}

\title[\bench: A Benchmark for Core Intent Identification in Personalized Question Answering]{\bench: A Benchmark for Core Intent Identification \\ in Personalized Question Answering}

\author{Jieyong Kim}
\affiliation{%
  \institution{Yonsei University}
  \city{Seoul}
  \country{Republic of Korea}}
\email{jieyong99@yonsei.ac.kr}

\author{Maryam Amirizaniani}
\affiliation{%
  \institution{University of Washington}
  \city{Seattle}
  \state{WA}
  \country{United States}}
\email{amaryam@uw.edu}

\author{Soojin Yoon}
\affiliation{%
  \institution{Yonsei University}
  \city{Seoul}
  \country{Republic of Korea}}
\email{soojiny@yonsei.ac.kr}

\author{Dongha Lee\textsuperscript{$\dagger$}}
\affiliation{%
  \institution{Yonsei University}
  \city{Seoul}
  \country{Republic of Korea}}
\email{donalee@yonsei.ac.kr}
\renewcommand{\shortauthors}{Jieyong Kim, Maryam Amirizaniani, Soojin Yoon, and Dongha Lee}

\begin{abstract}
  Intent identification serves as the foundation for generating appropriate responses in personalized question answering (PQA).
However, existing benchmarks evaluate only response quality or retrieval performance without directly measuring intent identification capabilities.
This gap is critical because without understanding which intents users prioritize, systems cannot generate responses satisfying individual information needs.
To address this, we introduce the concept of \textit{core intents}: intents users prioritize when selecting answers to satisfy their information needs.
To evaluate these core intents, we propose \textbf{\bench}, a benchmark for core \underline{\textbf{I}}ntent identification in \underline{\textbf{P}}ersonalized \underline{\textbf{Q}}uestion \underline{\textbf{A}}nswering.
Since users do not explicitly state their prioritized intents, we derive core intents from observable behavior patterns in answer selection, grounded in bounded rationality, where users satisfice by choosing answers meeting their acceptance thresholds.
We construct a dataset with various domains through systematic filtering, LLM-based annotation, and rigorous quality control combining automated verification with human validation.
Experimental evaluations across state-of-the-art language models reveal that current systems struggle with core intent identification in personalized contexts.
Models fail to identify core intents from user histories, with performance degrading as question complexity increases.
\href{https://github.com/jieyong99/IPQA}{\textcolor{blue}{[REPOSITORY]}}
\blfootnote{\textsuperscript{$\dagger$}Corresponding author.}%
\end{abstract}




\maketitle

\section{Introduction}
\label{sec:intro}

\begin{figure}[t]
  \centering
  \includegraphics[width=\columnwidth]{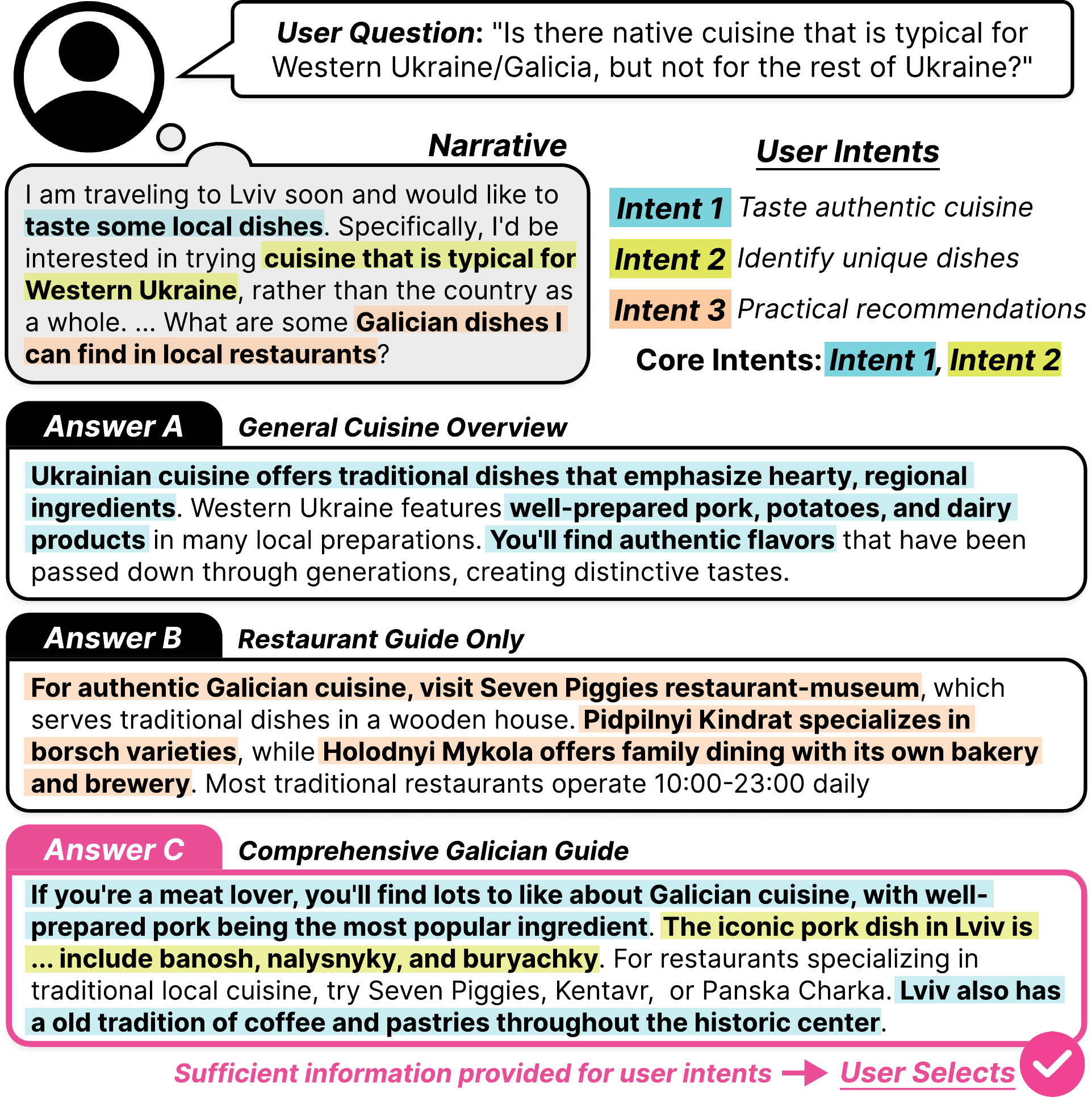}
  \caption{In information seeking scenarios, users ask questions with multiple intents and select answers that align with their prioritized intents. We define these as the core intents.}
  \label{fig:motiv}
\end{figure}

Intent understanding constitutes a fundamental capability that enables systems to interpret the underlying motivations driving user utterances.
Various intent identification benchmarks evaluate this capability, from single-intent classification within predefined taxonomies~\cite{larson-etal-2019-evaluation,casanueva-etal-2020-efficient,coucke2018snips} to open-world detection of unseen intents~\cite{zhang2021deep,zhang-etal-2021-textoir} and multi-intent identification~\cite{qin-etal-2020-agif,yoon-etal-2024-blendx}.
However, these benchmarks assume fixed intent sets for each utterance, which becomes inadequate in personalized question answering (\pqa) where identical questions express different intents depending on individual user backgrounds and information-seeking goals.

In \pqa, users seek information shaped by personal contexts~\cite{quarteroni-2010-personalized}, making intent understanding critical as the same question may require fundamentally different responses.
As illustrated in Figure~\ref{fig:motiv}, a question about regional cuisine may seek authentic local flavors and unique dishes, while another user with the same question might instead seek budget-friendly options or dietary accommodations.
However, existing \pqa benchmarks~\cite{salemi2025lampqabenchmarkpersonalizedlongform,10.1145/3589335.3651445,du-etal-2024-perltqa,10.1145/3711896.3737385} focus solely on response generation quality or retrieval performance, leaving intent identification in \pqa scenarios underexplored.

Beyond this absence of evaluation frameworks, identifying and evaluating intents in \pqa presents fundamental challenges.
Prior work on open intent detection has observed that users express multiple intents within a single utterance, with these intents varying in importance~\cite{10.1109/TASLP.2023.3265203}.
In personalized contexts, this complexity intensifies as both the intents expressed and their relative importance depend on individual user backgrounds and information-seeking goals.
These characteristics make intent identification difficult to predict and even more challenging to evaluate—without understanding which intents users prioritize, evaluation cannot assess whether systems recognize the important motivations that drive information need satisfaction.

Evaluating whether systems successfully identify intents that users prioritize requires verifiable ground truth that reflects user priorities without subjective judgment.
Such ground truth can be derived from observable user behavior in information seeking scenarios.
Bounded rationality~\cite{simon1955behavioral, agosto2002bounded} suggests that individuals operating under cognitive and resource constraints select solutions meeting acceptable thresholds rather than pursuing exhaustive optimization.
This behavioral pattern has been empirically validated through user studies examining real user cognition and selection behavior in information-seeking contexts~\cite{agosto2002bounded, chen2012understanding, prabha2007enough}: users select answers that satisfy their minimal information needs rather than pursuing exhaustive search, and their answer selections reliably reflect their prioritized motivations.

Based on this empirically validated behavioral pattern, we introduce the concept of \textit{core intents}—intents that align with information in selected answers, representing motivations users demonstrably prioritized when establishing their acceptance threshold.
As illustrated in Figure~\ref{fig:motiv}, a user initially poses a question with three intents, while the selected answer addresses only two of these intents—indicating that the user's minimum acceptance threshold for information need satisfaction prioritizes these two intents.
This approach provides a pragmatic evaluation framework grounded in observable behavior rather than subjective importance judgments.
While satisfying every expressed intent represents an ideal scenario, real-world information seeking involves prioritization under constraints. 
Core intents capture this reality: they represent the intents users demonstrably considered sufficient for satisfaction, as evidenced by answer selection. 
Though this may not capture every possible user motivation, it provides the most objective ground truth available without requiring subjective importance judgments.

To this end, we propose \textbf{\bench}, a benchmark for evaluating core \underline{\textbf{I}}ntent identification in \underline{\textbf{P}}ersonalized \underline{\textbf{Q}}uestion \underline{\textbf{A}}nswering.
Evaluating core intent identification capabilities requires datasets with user questions and their core intents, but collecting such data is challenging as users rarely state their core intents explicitly.
Following prior work on \pqa~\cite{salemi2025lampqabenchmarkpersonalizedlongform}, we utilize a dataset constructed from community question answering platforms where users provide not only questions but also detailed narratives explaining their underlying motivations and intentions about the question, alongside selected answers demonstrating satisfaction.
These narratives and selected answers serve as source data from which core intents can be derived, enabling systematic collection with complete user posting histories.
Construction proceeds through systematic filtering to ensure personalization requirements, followed by LLM-based intent annotation and rigorous quality control combining automated verification with human validation.
To evaluate core intent identification performance, we design \eval, an evaluation framework using an LLM-based evaluator to compare system predictions against annotated core intents, with meta-evaluation demonstrating strong alignment with human judgment.

The main contributions are summarized as follows:
\begin{itemize}
    \item Introduction of the concept of \textit{core intents} in \pqa, representing intents users prioritize when selecting answers, grounded in bounded rationality and observable user behavior.
    \item Construction of \bench, a benchmark comprising a dataset with rigorous quality control and an evaluation framework validated against human judgment.
    \item Experimental findings revealing that current language models struggle with core intent identification, failing to extract intent patterns from user histories with performance degrading as question complexity increases.
\end{itemize}

\section{IPQA Benchmark}
\label{sec:benchmark}
\begin{figure*}[t]
  \centering
  \includegraphics[width=0.99\textwidth]{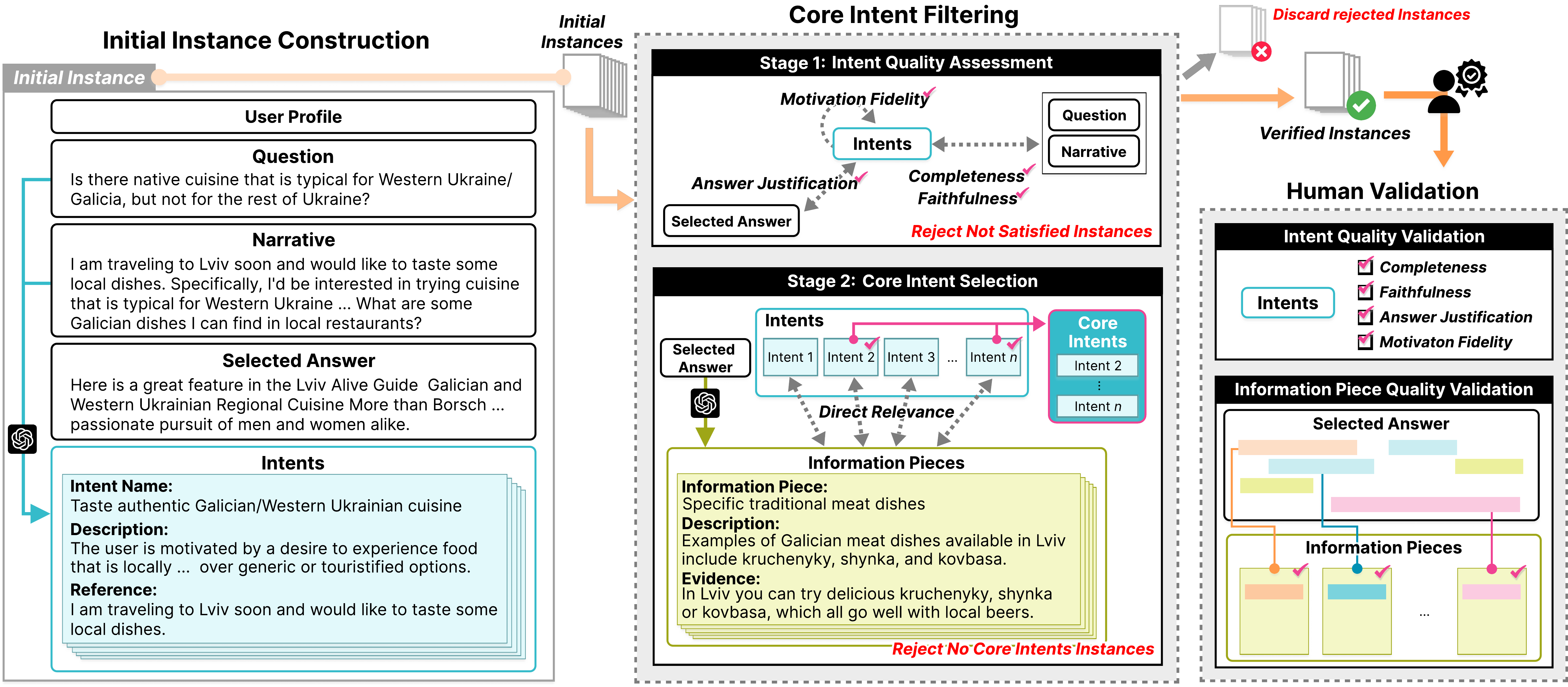}
  \caption{Overview of the \bench dataset construction pipeline: data collection from cQA dataset, LLM-based intent annotation, and quality control through LLM verification and human validation.}
  \label{fig:main_figure} 
\end{figure*}

In this section, we present \bench, a benchmark designed to evaluate core intent identification capabilities in personalized question answering (\pqa) scenarios.
The benchmark evaluates systems' ability to identify core intents in \pqa.
These core intents represent the specific intents that users prioritized when they select answers.
Construction of \bench proceeds through four steps: 
collecting \pqa instances and generating initial intent annotations (Section~\ref{subsec:initial_instance_construction}), 
verifying data quality and filtering core intents (Section~\ref{subsec:core_intent_filtering}), 
and validating annotations through human evaluation (Section~\ref{subsec:manual_validation}), 
and establishing evaluation metrics for core intent identification performance (Section~\ref{subsec:evaluation_framework}).
Figure~\ref{fig:main_figure} shows the overall process of our benchmark dataset construction.
All prompts throughout the pipeline were designed following established practices in benchmark construction~\cite{salemi2025lampqabenchmarkpersonalizedlongform,10.5555/3737916.3738608,seo-etal-2025-mt} with minimal necessary instructions, and are provided in the source code repository.

\subsection{Task Formulation}
\label{subsec:task_formulation}
The goal of the core intent identification task in \pqa is to predict core intents $I_c$ for a given question $q$ posed by user $u$.
To enable personalization, the system receives user profile $P_u = \{(q_i, s_i)\}_{i=1}^{|P_u|}$ containing historical questions $q_i$ paired with source information $s_i = (n_i, a_i)$ where $n_i$ is the narrative and $a_i$ is the selected answer, following previous personalization studies~\cite{salemi2023lamp,kumar2024longlampbenchmarkpersonalizedlongform,salemi2025lampqabenchmarkpersonalizedlongform,kim2025rpmreasoninglevelpersonalizationblackbox,10.1145/3726302.3730055}.
Since users do not explicitly articulate their prioritized intents in real-world scenarios, core intents must be inferred from $s$, reflecting realistic task conditions.
The system $\mathcal{M}$ leverages these historical records to predict core intents for the current question: $\hat{I}_c = \mathcal{M}(q, P_u)$.
The evaluation framework \eval compares predicted core intents $\hat{I}_c$ with ground truth $I_c$, producing intent identification scores: $S_{intent} = \eval(\hat{I}_c, I_c)$.
\rev{In this work, we distinguish three related concepts: \textit{intents} refer to individual motivations driving specific questions (e.g., ``taste authentic local cuisine''), \textit{topics} refer to subject areas that remain constant across askers, and \textit{user preferences} capture general tendencies across a user's history that shape but do not determine intents for a given question. Our benchmark focuses on identifying \textit{core intents} that users prioritize when selecting answers.}

\subsection{Initial Instance Construction}
\label{subsec:initial_instance_construction}

\subsubsection{Data Collection}
\label{subsubsec:data_collection}

Collecting real-world personalized question answering data—where users pose questions and express intents—presents substantial practical challenges.
Following established practices in \pqa research~\cite{salemi2025lampqabenchmarkpersonalizedlongform}, we utilize the SE-PQA dataset \cite{10.1145/3589335.3651445}, \rev{the only large-scale resource providing (Question, Narrative, Selected Answer) triplets necessary for our task.}
This dataset is constructed from community question answering (cQA) platforms where users provide questions with narratives containing detailed descriptions of the questions and select answers demonstrating resolution (Figure~\ref{fig:main_figure} left; detailed instances in Table~\ref{tbl:case_study}).
This aligns with our benchmark requirements: narratives reveal user motivations, while answer selections reflect satisficing behavior—users choose answers meeting minimum information needs rather than pursuing exhaustive optimization~\cite{simon1955behavioral, agosto2002bounded}.
These narratives and selected answers serve as source data for deriving core intents, while rich user posting histories enable personalization settings.

However, not all cQA instances require personalization—some questions yield identical answers regardless of who asks.
To filter such factoid questions, we select SE-PQA instances that have been verified as requiring personalization by LaMP-QA~\cite{salemi2025lampqabenchmarkpersonalizedlongform}. \rev{All utilized instances contain accepted answers explicitly selected by the original question askers, ensuring that answer selection reflects the asker's own judgment rather than community voting or third-party curation.}
This verification employed two-stage validation combining LLM-based filtering with human verification to ensure these instances genuinely require user-specific context.
Each instance contains question $q$, narrative $n$, and selected answer $a$.
For each user, we collect all qualifying instances and construct profiles chronologically: the most recent instance serves as the target question for evaluation, while all preceding instances form the user profile $P_u = \{(q_i, s_i)\}_{i=1}^{|P_u|}$ where $s_i = (n_i, a_i)$.
Since users do not explicitly state their core intents, these source data enable intent inference through observable behavior patterns.
This yields initial instances structured as $(q, s, P_u)$, providing the foundation for subsequent intent annotation.

\subsubsection{Initial Intent Generation}
\label{subsubsec:initial_intent_generation}

Establishing intent ground truth requires annotation, as these intents do not exist explicitly in the original data and must be generated.
Manual annotation at dataset scale presents practical challenges due to prohibitive costs and difficulty maintaining consistency across annotators~\cite{klie-etal-2024-analyzing}.
To address these limitations, the annotation process employs automated generation leveraging LLM capabilities, following established practices in benchmark construction~\cite{seo-etal-2025-mt,salemi2025lampqabenchmarkpersonalizedlongform,heo2025largelanguagemodelseffective}.
The framework utilizes GPT-5-Mini\footnote{\url{https://platform.openai.com/docs/models/gpt-5-mini}} for initial intent generation, followed by rigorous quality control detailed in Section~\ref{subsec:core_intent_filtering} and~\ref{subsec:manual_validation}.

\rev{While initial instances contain both narratives and selected answers, intent generation uses only narratives to capture pre-selection motivations; selected answers are later used in core intent filtering (Section 2.3.2) to identify which intents users prioritized. This separation ensures that generated intents reflect pre-selection motivations rather than post-hoc rationalizations of the chosen answer.}

Intent generation produces comprehensive intent sets from narratives $n$.
The model generates multiple atomic intents through prompting, where each intent represents a distinct motivation reflecting the multi-intent nature of personalized questions.
Actual annotated instance is provided in Appendix~\ref{apdx:case_study} for reference.
Each generated intent contains three components: \textit{intent name} (brief label), \textit{description} (detailed explanation), and \textit{reference} (supporting text from $n$).
This process captures initial motivations users hold when posing questions, as expressed in their narratives.

\subsection{Core Intent Filtering}
\label{subsec:core_intent_filtering}

Initial intent generation requires verification to ensure annotation quality and identification of which intents constitute core intents (motivations users prioritized when selecting answers).
Following established \textit{Generate-then-Filter} methodology~\cite{heo2025largelanguagemodelseffective,salemi2025lampqabenchmarkpersonalizedlongform,seo-etal-2025-mt}, our pipeline operates through two sequential stages using an LLM evaluator, where each stage produces verifiable intermediate outputs validated independently before proceeding.
This modular design prevents error propagation: subsequent filtering catches low-quality outputs rather than amplifying them.

The first stage filters instances based on multiple quality criteria applied to initial intents.
The second stage identifies core intents from quality-verified intents through fine-grained alignment assessment, where information pieces extracted from selected answers are mapped to intents, with successfully mapped intents constituting the core intent set.
This process enables filtering low-quality instances and annotating core intents without requiring explicit user articulation.
Subsequent human validation verifies the reliability of this automated filtering (Section~\ref{subsec:manual_validation}).

\subsubsection{Intent Quality Assessment}
\label{subsubsec:intent_quality_assessment}

To ensure dataset quality, generated initial intents undergo filtering using LLM as a self-verifier.
Each generated intent is evaluated against four binary criteria. \rev{Following recent benchmarks~\cite{10.5555/3737916.3738608, seo-etal-2025-mt}, \textit{completeness} checks coverage of all narrative motivations and \textit{faithfulness} excludes extraneous content; as task-specific extensions, \textit{motivational fidelity} requires genuine underlying reasons and \textit{answer justification} ensures a sufficient basis for later alignment assessment (Section 2.3.2).}
Instances containing any intent that fails a criterion are removed, yielding high quality instances for subsequent core intent selection.

\subsubsection{Core Intent Selection}
\label{subsubsec:core_intent_selection}

Core intents represent motivations users prioritized when selecting answers in \pqa.
Quality-verified instances undergo further processing to identify these core intents through alignment assessment with selected answers, using LLM as an evaluator.
For fine-grained comparison, information pieces are extracted from selected answers $a$ through atomic decomposition, following established practices~\cite{salemi-etal-2025-expert}.
To ensure meaningful alignment assessment, each piece must represent a coherent, self-contained unit of information, which we verify through \textit{Information Piece Quality} evaluation.
For each validated information piece, the LLM evaluator assesses \textit{Direct Relevance} to identify which intents are directly addressed by the answer content.
Individual intents may align with multiple information pieces, reflecting their relevance to various information of the answer.
Intents successfully mapped to at least one information piece constitute core intents for that instance, while instances where no intents map to any information piece undergo filtering as the absence of core intents prevents evaluation.
These annotated core intents serve as verifiable ground truth references for evaluating model-predicted core intents.

\subsection{Human Validation}
\label{subsec:manual_validation}
To verify automated annotation quality, human evaluators assess 400 instances sampled with stratification across domains \rev{(95\% CI $\pm$3.4\%; Entertainment 36\%, Living 33\%, Social 31\%)} using the same criteria applied in Section~\ref{subsec:core_intent_filtering}, along with an additional evaluation of information piece extraction quality.
\rev{Annotators evaluate all generated intents and all extracted information pieces for each sampled instance, assessing annotation quality across the full pipeline rather than only for selected core intents.}
\rev{Two trained CS/AI graduate students served as independent annotators,} evaluating each instance on five-point Likert scales with results presented in Table~\ref{tab:human_eval}. \rev{\textit{Perfect Agreement} denotes the percentage of instances where both annotators assign identical scores on the five-point Likert scale, while \textit{Both-5 Rate} denotes the percentage where both annotators assign the maximum score of 5.}
Human evaluation demonstrates high annotation quality, \rev{with perfect agreement averaging 89.8\% and both-5 rate averaging 93.4\%.}
These results validate that both intent annotations and information piece extractions maintain high quality standards aligned with human judgment.
\rev{Detailed annotation guidelines and training procedures are provided in Appendix~\ref{apdx:annotation}.}
Final dataset statistics are presented in Table~\ref{tab:dataset_statistics}, with detailed breakdowns in Section~\ref{apdx:benchmark_details}.

\begin{table}[t]
\small
\centering
\caption{Human evaluation results.}
\vspace{-5pt}
\resizebox{\columnwidth}{!}{
\begin{tabular}{lcc}
    \toprule
    \textbf{Data Quality Criteria} & \textbf{Perfect Agreement} & \textbf{Both-5 Rate} \\ 
    \midrule[\heavyrulewidth]
    Intent-Narrative Completeness & 88.5 & 96.0 \\
    Intent-Narrative Faithfulness & 95.0 & 95.5 \\
    \midrule
    Intent Motivational Fidelity & 85.9 & 89.4 \\
    Intent-Answer Justification & 90.2 & 97.7 \\
    Intent-Information Piece Direct Relevance & 92.5 & 95.0 \\
    Information Piece Quality & 86.5 & 86.5 \\
    \midrule
    \textbf{Average} & \textbf{89.8} & \textbf{93.4} \\
    \bottomrule
\end{tabular}
}
\label{tab:human_eval}
\end{table}

\subsection{\eval}
\label{subsec:evaluation_framework}

To evaluate core intent identification performance, we design \eval, an evaluation framework comparing model-predicted core intents against annotated ground truth core intents.
The framework employs LLM-based matching to assess alignment between predicted intents and ground truth intents, capturing whether they reflect the same underlying motivations.
Each predicted atomic intent is compared against all ground truth atomic intents to identify alignments.
However, direct one-to-one comparison between all predicted and ground truth intent pairs incurs prohibitive computational costs.
Therefore, this one-to-many comparison approach~\cite{zhang2020bertscoreevaluatingtextgeneration, salemi-etal-2025-expert} efficiently compares each predicted intent against all ground truth intents in a single evaluation call to find the best alignment.
This reduces computational complexity from $O(|I_c| \times |\hat{I}_c|)$ to $O(|\hat{I}_c|)$, where $|I_c|$ and $|\hat{I}_c|$ denote the number of ground truth and predicted core intents respectively.
When multiple predicted intents align with the same ground truth intent, that ground truth intent is counted as correctly identified only once in recall calculation.
Based on these alignments, we compute precision, recall, and F1 to measure intent identification performance.

\begin{table}[t]
\small
\centering
\caption{Statistics of the \bench dataset.}
\vspace{-5pt}
\resizebox{\columnwidth}{!}{
\begin{tabular}{lccc}
    \toprule
    \textbf{Properties} & \textbf{Entertainment} & \textbf{Living} & \textbf{Social} \\
    \midrule[\heavyrulewidth] 
    \# of train set                   & 1785             & 1360            & 1510            \\
    \# of validation set              & 610              & 444             & 476             \\
    \# of test set                    & 610              & 424             & 511             \\
    \midrule[\heavyrulewidth]
    Avg. profile size       & 91.08 $\pm$ 112.54 & 80.93 $\pm$ 106.05 & 83.07 $\pm$ 107.28 \\
    Avg. core intent        & 3.42 $\pm$ 0.96    & 3.81 $\pm$ 0.98    & 3.77 $\pm$ 0.95    \\
    Avg. info. piece             & 6.97 $\pm$ 3.33    & 7.94 $\pm$ 3.23    & 7.89 $\pm$ 3.26    \\
    Avg. pieces per intent & 2.33 $\pm$ 1.71    & 2.33 $\pm$ 1.60    & 2.35 $\pm$ 1.62    \\
    \midrule[\heavyrulewidth]
    $1\leq|I_c|\leq2$          & 469 (15.61\%)        & 179 (8.03\%)        & 200 (8.01\%)       \\
    $3\leq|I_c|\leq4$        & 2175 (72.38\%)     & 1542 (69.21\%)      & 1780 (71.29\%)      \\
    $5\leq|I_c|$    & 361 (12.01\%)     & 507 (22.76\%)     & 517 (20.70\%)     \\
    \midrule[\heavyrulewidth]
    Total                   & 3005              & 2228              & 2497              \\
    \bottomrule
\end{tabular}
}
\label{tab:dataset_statistics}
\end{table}

\section{Experiments}
\label{sec:exp}

In this section, we conduct experiments to examine how well LLMs identify core intents in personalized contexts using \bench.

\subsection{Experimental Setup}

\subsubsection{Baselines}
We design four baselines varying the user-specific context provided to the model.

\noindent
\textbf{\nopers.} 
Core intent identification in \pqa requires user context to understand individual intent priorities.
To establish baseline performance without personalization, the model receives only $q$ without $P_u$, generating intents based solely on the question.
This configuration measures core intent identification capabilities in the absence of user-specific context.

\noindent
\textbf{\ranicl.} 
Profiles may contain generic patterns or domain-specific information that could improve performance regardless of user-specific personalization.
To isolate the effect of user-specific patterns, the model receives $(q_i, s_i)$ pairs randomly selected from a random $P$ as in-context examples.
This configuration verifies whether observed performance gains stem from genuine personalization rather than generic historical information.

\noindent
\textbf{\rawrag.} 
Core intent identification performance may improve when models access user-specific historical patterns that reveal individual preferences and intent priorities.
To test this hypothesis, the model receives $(q_i, s_i)$ pairs retrieved from $P_u$ based on semantic similarity between $q$ and $s_i$.
This configuration tests whether models can capture user-specific historical patterns from past $s_i$ and leverage them for personalized intent generation on $q$.

\noindent
\textbf{\intrag.}
As in real-world scenarios where users do not explicitly articulate their core intents, systems must infer them from $s_i$ during inference.
However, models may struggle to extract intent patterns directly from raw $s_i$ in $P_u$.
To test whether preprocessing alleviates this challenge, each historical item undergoes automatic intent derivation before model inference.
Intents are derived from $s_i$ based on alignment with $a_i$, using GPT-4.1-Mini with 3-shot prompting from training examples, following the procedure in Section~\ref{subsec:core_intent_filtering} without human validation.
During inference, the model receives $(q_i, I_{i})$ pairs for the same items retrieved as \rawrag, but with derived intents $I_{i}$ instead of raw $s_i$.
This configuration tests whether preprocessed intents enable more effective personalization than raw $s_i$.

\subsubsection{Implementation Details}

We evaluate multiple LLMs: Llama-3.1-8B-It~\cite{dubey2024llama}, Qwen2.5-7B-It~\cite{hui2024qwen2}, Gemma-3-12B-It~\cite{team2025gemma}, Mistral-Small-24B-It-2501\footnote{\url{https://huggingface.co/mistralai/Mistral-Small-24B-Instruct-2501}}, Qwen2.5-32B-It~\cite{hui2024qwen2}, and GPT-4o-Mini\footnote{\url{https://platform.openai.com/docs/models/gpt-4o-mini}}.
For retrieval, Qwen3-Embedding-8B~\cite{zhang2025qwen3} encodes questions and retrieves the top-$k=30$ profile items based on cosine similarity to provide rich personalized context as default.
We employ GPT-4.1-Mini\footnote{\url{https://platform.openai.com/docs/models/gpt-4.1-mini}} as the evaluator across all experimental conditions through \eval considering performance and cost.
All experiments utilize vLLM~\cite{kwon2023efficient} for inference with temperature 0.0 and no output length constraints.
Profile items exceeding 1,024 tokens are excluded during retrieval.
Experiments are conducted on a system equipped with 4 NVIDIA A6000 GPUs (48GB VRAM each), 1024GB system RAM, and Intel Xeon Gold 6526Y CPU.
Due to space constraints, all prompts are provided in the source code repository.

\subsection{Meta-Evaluation}
\label{subsec:meta_evaluation}

To ensure reliable quality assessment, automated evaluation metrics should demonstrate strong alignment with human judgment.
Following established meta-evaluation methodology~\cite{seo-etal-2025-mt,10.5555/3737916.3738608}, we validate \eval's reliability by measuring its correlation with human preferences compared to existing automated metrics.
The meta-evaluation dataset comprises 100 randomly sampled instances from the test set, where each instance contains two different intent responses generated using GPT-4o-Mini with temperature 1.0 under identical \rawrag configurations.
\rev{Two trained AI researchers} evaluate each pair through pairwise comparison on two dimensions: completeness, which assesses whether generated intents capture all essential motivations (corresponding to recall in our metrics), and faithfulness, which evaluates whether intents preserve original context without hallucination (corresponding to precision). \rev{These two dimensions map directly to recall (completeness) and precision (faithfulness), while motivational fidelity and answer justification from Section 2.3.1 serve as binary construction filters unsuitable for pairwise comparison.}
Annotation details are provided in Appendix~\ref{apdx:annotation}.

\begin{table}[t]
\small
\centering
\caption{Meta-Evaluation results.}
\vspace{-5pt}
\begin{tabular}{lcc}
    \toprule
    \textbf{Evaluation Metric} &  Completeness & Faithfulness \\ 
    \midrule[\heavyrulewidth]
    SacreBLEU~\cite{post-2018-call}                          & 31.37        & 27.03        \\
    ROUGE-L~\cite{lin-hovy-2003-automatic}                            & 34.11        & 25.41        \\
    METEOR~\cite{banerjee-lavie-2005-meteor}                             & 37.78        & 34.42        \\
    BERTScore~\cite{zhang2020bertscoreevaluatingtextgeneration}                          & 38.00           & 36.39        \\
    G-Eval~\cite{liu-etal-2023-g}                             & \underline{48.16}        & \underline{56.38}        \\
    \midrule
    \eval                          & \textbf{60.59}        & \textbf{63.68}        \\
    \midrule
    Inter-Human Correlation            & 78.57        & 80.32        \\ 
    \bottomrule
\end{tabular}
\label{tab:metaeval_intent}
\end{table}

Since human preference labels can be seen as the score difference of a response pair: $h_i = H(r_i^2) - H(r_i^1) \in \{-1, 0, 1\}$, with an evaluation model $E$, we compute a normalized score difference as $e_i = f(E(r_i^2) - E(r_i^1)) \in [-1, 1]$, where $f$ is a linear normalization function.
Meta-evaluation measures the \rev{Kendall's $\tau$} correlation between $h_i$ and $e_i$ across all 100 instances, together with the \rev{Kendall's $\tau$} correlation between $h_i$ and $h_i'$ from two annotators as the upper bound.
\rev{For each baseline metric, scores are computed for both responses and normalized as for $e_i$, with the completeness and faithfulness columns reporting correlations against coverage and accuracy judgments, respectively.}

Table~\ref{tab:metaeval_intent} demonstrates that \eval achieves the highest correlation with human preferences across both dimensions, substantially outperforming \rev{metrics spanning diverse evaluation paradigms: lexical overlap (SacreBLEU~\cite{post-2018-call}, ROUGE-L~\cite{lin-hovy-2003-automatic}), semantic matching (METEOR~\cite{banerjee-lavie-2005-meteor}, BERTScore~\cite{zhang2020bertscoreevaluatingtextgeneration}), and LLM-based evaluation (G-Eval~\cite{liu-etal-2023-g} using GPT-4.1-Mini with 1-5 Likert scales)}.

Beyond human alignment, we verify evaluation robustness across model families to address potential circularity concerns.
Using outputs from all models under \intrag configuration in Table~\ref{tbl:main}, we test Qwen2.5-32B-It and Mistral-Small-24B as independent evaluators and measure their agreement with our default evaluator.
Results show strong cross-model consistency: Qwen achieved 75.1\% and Mistral 81.1\% pairwise agreement with GPT-4.1-Mini, while majority voting across all three evaluators reached 96.9\% agreement.
This consistency, combined with strong human correlation (Table~\ref{tab:metaeval_intent}), confirms that IPQA-Eval captures genuine quality dimensions rather than model-specific artifacts. \rev{We note that cross-model consistency demonstrates evaluation robustness but does not fully resolve the circularity of using LLMs for both construction and evaluation, which we discuss in Section~\ref{sec:limitations}.}

\begin{table*}[t]
\centering
\small
\setlength{\tabcolsep}{2pt} 
\caption{Performance for core intent identification on the test set (*: significant vs. other baselines, paired t-test, $p$ < 0.05).}
\vspace{-5pt}
\begin{tabular}{ll ccc ccc ccc ccc}
    \toprule
    \multirow{2.5}{*}{\textbf{Model}} & \multirow{2.5}{*}{\textbf{Method}} & \multicolumn{3}{c}{\textbf{Entertainment}} & \multicolumn{3}{c}{\textbf{Living}} & \multicolumn{3}{c}{\textbf{Social}} & \multicolumn{3}{c}{\textbf{Average (Macro)}} \\
    \cmidrule(lr){3-5} \cmidrule(lr){6-8} \cmidrule(lr){9-11} \cmidrule(lr){12-14}
    & & Precision & Recall & F1 & Precision & Recall & F1 & Precision & Recall & F1 & Precision & Recall & F1 \\
    \midrule
    \multirow{4.3}{*}{Llama-3.1-8B-It}
    & \nopers 
    & 0.4281 & 0.3757 & 0.3798 & \underline{0.4901} & \textbf{0.4226} & \underline{0.4290} & \underline{0.4705} & \underline{0.4069} & \underline{0.4136} & \underline{0.4629} & \underline{0.4017} & \underline{0.4075} \\
    & \ranicl 
    & 0.4185 & 0.3315 & 0.3519 & 0.4887 & 0.3803 & 0.4073 & 0.4619 & 0.3773 & 0.3928 & 0.4564 & 0.3630 & 0.3840 \\
    & \rawrag 
    & \underline{0.4605} & \underline{0.3796} & \underline{0.3932} & 0.4788 & \underline{0.4006} & 0.4154 & 0.4354 & 0.3697 & 0.3823 & 0.4582 & 0.3830 & 0.3969 \\
    \cmidrule(lr){2-14}
    & \intrag 
    & \textbf{0.5643*} & \textbf{0.4228*} & \textbf{0.4608*} & \textbf{0.5500*} & 0.3991 & \textbf{0.4410*} & \textbf{0.4747} & \textbf{0.4125*} & \textbf{0.4146} & \textbf{0.5297*} & \textbf{0.4115*} & \textbf{0.4388*} \\
    \midrule[\heavyrulewidth]
    \multirow{4.3}{*}{Qwen2.5-7B-It}
    & \nopers 
    & \underline{0.5574} & 0.3806 & \underline{0.4306} & \underline{0.6294} & 0.4186 & \underline{0.4752} & \textbf{0.5928*} & 0.3905 & \underline{0.4426} & \underline{0.5932} & 0.3966 & \underline{0.4495} \\
    & \ranicl 
    & 0.4569 & 0.3464 & 0.3735 & 0.5288 & 0.4045 & 0.4335 & 0.5098 & 0.4000 & 0.4267 & 0.4985 & 0.3836 & 0.4112 \\
    & \rawrag 
    & 0.4971 & \underline{0.4024} & 0.4234 & 0.5353 & \underline{0.4251} & 0.4513 & 0.5089 & \underline{0.4081} & 0.4314 & 0.5138 & \underline{0.4119} & 0.4354 \\
    \cmidrule(lr){2-14}
    & \intrag 
    & \textbf{0.6458*} & \textbf{0.4866*} & \textbf{0.5282*} & \textbf{0.6316} & \textbf{0.4723*} & \textbf{0.5172*} & \underline{0.5557} & \textbf{0.4475*} & \textbf{0.4694*} & \textbf{0.6110*} & \textbf{0.4688*} & \textbf{0.5049*} \\
    \midrule[\heavyrulewidth]
    \multirow{4.3}{*}{Gemma3-12B-It}
    & \nopers 
    & 0.4827 & 0.4437 & 0.4439 & \underline{0.5713} & \textbf{0.4832} & \underline{0.5025} & \underline{0.5403} & \textbf{0.4673} & \underline{0.4768} & 0.5314 & \underline{0.4647} & \underline{0.4744} \\
    & \ranicl 
    & \underline{0.5403} & 0.4108 & 0.4425 & 0.5631 & 0.4312 & 0.4667 & 0.5364 & 0.4170 & 0.4471 & \underline{0.5466} & 0.4197 & 0.4521 \\
    & \rawrag 
    & 0.5338 & \underline{0.4566} & \underline{0.4688} & 0.5186 & 0.4537 & 0.4615 & 0.5215 & 0.4362 & 0.4520 & 0.5246 & 0.4485 & 0.4608 \\
    \cmidrule(lr){2-14}
    & \intrag 
   & \textbf{0.6125*} & \textbf{0.4875*} & \textbf{0.5177*} & \textbf{0.5994*} & \underline{0.4783} & \textbf{0.5066} & \textbf{0.5506*} & \underline{0.4666} & \textbf{0.4809} & \textbf{0.5875*} & \textbf{0.4775*} & \textbf{0.5017*} \\
    \midrule[\heavyrulewidth]
    \multirow{4.3}{*}{Mistral-Small-24B}
    & \nopers 
    & \underline{0.5294} & 0.4081 & 0.4374 & \underline{0.5937} & \underline{0.4559} & \underline{0.4912} & \textbf{0.5521} & 0.4204 & \underline{0.4520} & \underline{0.5584} & \underline{0.4281} & \underline{0.4602} \\
    & \ranicl 
    & 0.4887 & 0.3980 & 0.4165 & 0.5315 & 0.4237 & 0.4517 & 0.5028 & 0.4116 & 0.4311 & 0.5077 & 0.4111 & 0.4331 \\
    & \rawrag 
    & 0.5234 & \underline{0.4212} & \underline{0.4452} & 0.5249 & 0.4219 & 0.4532 & 0.5030 & \underline{0.4247} & 0.4396 & 0.5171 & 0.4226 & 0.4460 \\
    \cmidrule(lr){2-14}
    & \intrag 
    & \textbf{0.6460*} & \textbf{0.4969*} & \textbf{0.5364*} & \textbf{0.6209} & \textbf{0.4841*} & \textbf{0.5180*} & \underline{0.5313} & \textbf{0.4532*} & \textbf{0.4670} & \textbf{0.5994*} & \textbf{0.4781*} & \textbf{0.5068*} \\
    \midrule[\heavyrulewidth]
    \multirow{4.3}{*}{Qwen2.5-32B-It}
    & \nopers 
    & 0.4782 & 0.4353 & 0.4320 & \textbf{0.6138*} & 0.4540 & \underline{0.4956} & \underline{0.5090} & 0.4387 & 0.4427 & \underline{0.5337} & 0.4427 & 0.4568 \\
    & \ranicl 
    & 0.4956 & 0.4254 & 0.4357 & 0.5192 & 0.4660 & 0.4700 & 0.4835 & 0.4247 & 0.4307 & 0.4994 & 0.4387 & 0.4454 \\
    & \rawrag 
    & \underline{0.5313} & \underline{0.4707} & \underline{0.4746} & 0.5244 & \underline{0.4871} & 0.4816 & 0.4998 & \underline{0.4393} & \underline{0.4470} & 0.5185 & \underline{0.4657} & \underline{0.4677} \\
    \cmidrule(lr){2-14}
    & \intrag 
    & \textbf{0.6088*} & \textbf{0.5277*} & \textbf{0.5388*} & \underline{0.5539} & \textbf{0.4978*} & \textbf{0.5016} & \textbf{0.5122} & \textbf{0.4777*} & \textbf{0.4712*} & \textbf{0.5583*} & \textbf{0.5011*} & \textbf{0.5039*} \\
    \midrule[\heavyrulewidth]
    \multirow{4.3}{*}{GPT-4o-Mini}
    & \nopers 
    & 0.4451 & 0.3869 & 0.3935 & \underline{0.5877} & \underline{0.4483} & \underline{0.4831} & \underline{0.5366} & \underline{0.4247} & \underline{0.4475} & \underline{0.5231} & \underline{0.4200} & \underline{0.4413} \\
    & \ranicl 
    & 0.4125 & 0.3477 & 0.3597 & 0.5584 & 0.4259 & 0.4599 & 0.4913 & 0.3933 & 0.4183 & 0.4874 & 0.3890 & 0.4126 \\
    & \rawrag 
    & \underline{0.4683} & \underline{0.3986} & \underline{0.4084} & 0.5259 & 0.4420 & 0.4605 & 0.4787 & 0.4092 & 0.4231 & 0.4910 & 0.4166 & 0.4307 \\
    \cmidrule(lr){2-14}
    & \intrag 
    & \textbf{0.6054*} & \textbf{0.4977*} & \textbf{0.5165*} & \textbf{0.6237*} & \textbf{0.5055*} & \textbf{0.5304*} & \textbf{0.5539} & \textbf{0.4877*} & \textbf{0.4930*} & \textbf{0.5943*} & \textbf{0.4970*} & \textbf{0.5133*} \\
    \bottomrule
\end{tabular}
\label{tbl:main}
\end{table*}

\subsection{Core Intent Identification Results}
\label{subsec:main_results}

\textbf{Models fail to extract intent patterns from raw source data, but preprocessed intents enable effective personalization.}
Table~\ref{tbl:main} presents core intent identification performance.
\ranicl consistently underperforms \nopers, confirming that personalization requires user-specific patterns rather than arbitrary historical context.
\rawrag also performs worse than \nopers despite providing complete, relevant user histories, revealing that models fail to extract intent-level patterns from unstructured post histories.
In contrast, \intrag demonstrates substantial improvements, confirming that user histories encode stable, recoverable intent patterns that models can leverage when preprocessed.

\medskip
\noindent\textbf{Entertainment domain benefits most from personalization.}
Social domain presents the most challenging setting, while Entertainment and Living achieve comparably higher performance.
However, Entertainment shows the largest improvement from \nopers to \intrag, as questions about entertainment (e.g., movies or games) are more sensitive to individual differences.
Despite these gains, even the best-performing configuration achieves only moderate performance, revealing that core intent identification in \pqa remains fundamentally challenging.

\begin{figure*}[t]
  \centering
  \includegraphics[width=0.99\textwidth]{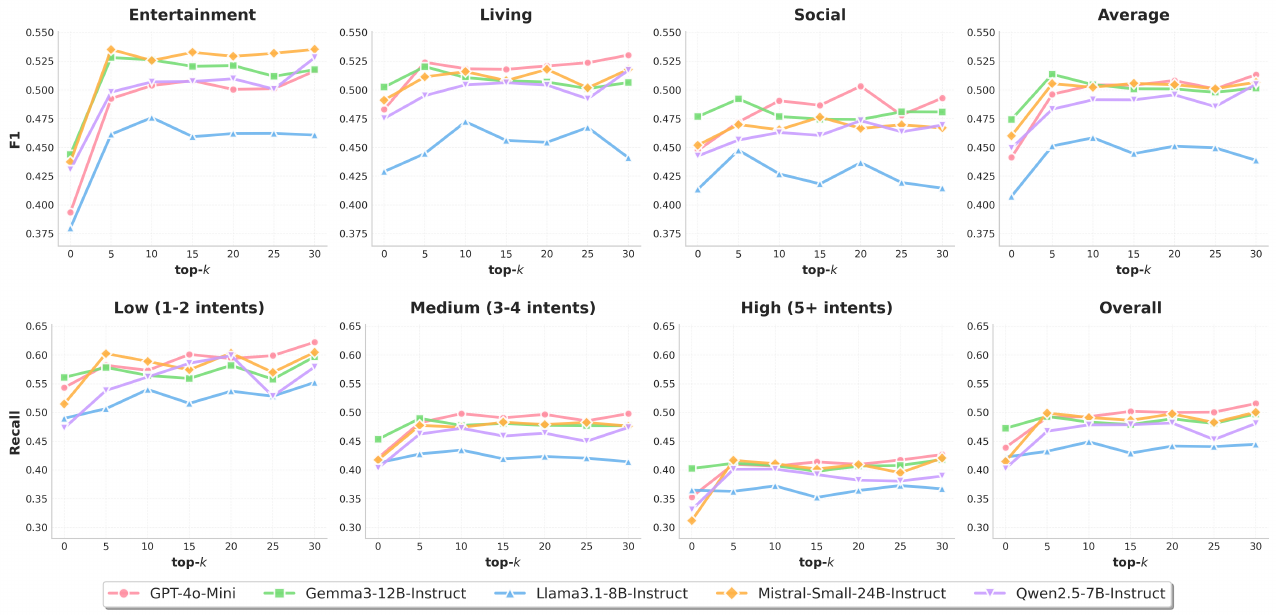}
  \caption{Core intent identification performance of \intrag with varying profile history sizes ($k$). Top: Results across different domains. Bottom: Results grouped by question complexity.}
  \label{fig:topk} 
\end{figure*}

\subsection{Impact of Profile History Size}
\label{subsec:topk_analysis}

\rev{\textbf{Initial profile items substantially improve core intent identification, with diminishing returns at larger history sizes.}}
Figure~\ref{fig:topk} (top) presents F1 scores across varying numbers of retrieved profile items under \intrag.
\rev{Most models show substantial improvements from $k=0$ to $k=5$, capturing the majority of performance gains with a small number of highly relevant profile items. Beyond $k=5$, improvements become marginal or inconsistent, suggesting that additional context may introduce noise or exceed models' effective context utilization capacity.}
All three domains benefit from increased history size, with Entertainment exhibiting the steepest improvement curves compared to Living and Social, consistent with the results observed in Section~\ref{subsec:main_results}.

\medskip
\rev{Optimal history size varies across models: Gemma3-12B-It peaks at $k=5$ then declines under noise, Llama-3.1-8B-It improves inconsistently under extended context, while larger models such as Qwen2.5-32B-It scale more stably.}
Despite these variations, even minimal personalization context ($k=5$) yields substantial gains over no personalization, confirming that user history utilization is critical for core intent identification. \rev{The diminishing returns beyond $k=5$ suggest that retrieval quality and context utilization efficiency may matter more than history volume, highlighting these as important directions for future improvement.}

\subsection{Question Complexity Analysis}
\label{subsec:complexity_analysis}

\textbf{Performance degrades as question complexity increases.}
Users typically express multiple intents within single questions in \pqa contexts (Table~\ref{tab:dataset_statistics}).
To examine how intent count affects identification, we group questions by core intents into Low (1-2), Medium (3-4), and High (5+) complexity levels.
Figure~\ref{fig:topk} (bottom) presents recall across these levels under \intrag with varying history sizes.
At $k=30$, where most models achieve best performance, all models show substantially lower recall for High-complexity questions than Low-complexity ones.

\medskip
\noindent\textbf{Historical context benefits all complexity levels but cannot close the gap.}
Average recall gains from $k=0$ to $k=30$ reach +18.9\% (Low), +14.2\% (Medium), and +16.6\% (High).
However, High-complexity questions maintain lower absolute recall even with extended history, confirming that identifying all core intents in multi-intent scenarios remains challenging.

\subsection{Intent-Answer Correlation Analysis}
\label{subsec:intent_answer_correlation}

Previous experiments established that core intent identification remains challenging for \pqa.
While accurate intent identification appears critical, its actual impact on answer generation, the goal of satisfying user information needs, remains unclear.
To measure this impact, we vary intent provision configurations while maintaining identical profile contexts using Profile RAG (Intents) \rev{as the base configuration given its strongest personalization performance (Table~\ref{tbl:main})}, then evaluate the quality of generated answers.
For this evaluation, we design \eval~(Answer), which extracts information pieces from generated answers and matches them against ground truth pieces from selected answers to compute precision, recall, and F1, following the same alignment approach as \eval~(Intent).
Meta-evaluation (Table~\ref{tbl:metaeval_answer}) demonstrates strong correlation with human judgment, validating this approach.

We evaluate answer generation under three intent provision configurations: (1) \textit{Predicted}---$\hat{I}_c$ predicted by models from $q$ and $P_u$, (2) \textit{Core Only}---exclusively ground truth $I_c$, and (3) \textit{Noisy}---$I_c$ mixed with random intents at 1:1 ratio.
During answer generation, all configurations share identical $k=30$ profile examples from $P_u$ as $(q_i, I_{c,i}, a_i)$ triplets, differing only in the intents provided for $q$.

\medskip
\noindent\textbf{Accurate core intent identification directly improves answer quality.}
Table~\ref{tab:intent_answer} shows the Core Only configuration achieves highest performance while Noisy maintains strong scores despite 50\% noise, whereas Predicted shows substantially lower performance.
This confirms that high recall of core intents benefits answer generation even when irrelevant intents are present, a pattern consistent across all domains.
However, even with all core intents provided, recall remains limited, as not all information in selected answers directly relates to core intents and generating specific details from abstract intents presents additional challenges.
Despite this, the substantial gap between Core Only and Predicted confirms that accurate core intent identification remains fundamental to generating answers that satisfy user information needs.

\medskip
\noindent\textbf{Correctly identifying individual core intents dramatically increases generation of corresponding information.}
To further investigate this relationship, we analyze how correctly identifying individual core intents affects generation of their corresponding information pieces.
We compute conditional probabilities: $P(\geq1|$\cmark$)$ measures the probability of generating at least one relevant information piece when the corresponding core intent was correctly identified, while $P(\geq1|$\xmark$)$ represents this probability when missed.
Similarly, $Acc|$\cmark~and $Acc|$\xmark~measure the accuracy of generating all associated information pieces, conditioned on whether that intent was correctly identified.
Table~\ref{tab:intent_answer_cond} reveals substantial gaps between correctly and incorrectly identified intents, with this pattern holding consistently across all models and domains.
The gap is particularly pronounced for generating at least one relevant piece, suggesting that correct intent identification substantially helps initiate relevant information generation, though producing all associated details in the answer remains challenging.

\medskip
\noindent\textbf{Qualitative analysis illustrates how intent errors redirect answers toward different topics.}
Table~\ref{tab:qualitative} presents a representative example where predicted intents capture only vague notions of balance while missing the user's prioritized themes of contrast and discovery.
The resulting answer discusses generic considerations entirely misaligned with what the user-selected answer addresses.
With ground truth intents, the generated answer shifts to directly address the user's actual priorities.

\section{Related Work}
\label{sec:relwork}

\subsection{Intent Understanding Benchmarks}
\label{subsec:intent_understanding}

Intent understanding benchmarks evaluate systems' ability to identify motivations underlying user utterances.
Early work establishes supervised classification frameworks using predefined intent taxonomies, with benchmarks such as CLINC150~\cite{larson-etal-2019-evaluation}, BANKING77~\cite{casanueva-etal-2020-efficient}, and SNIPS~\cite{coucke2018snips} providing labeled categories for task-oriented dialogue systems.
However, fixed taxonomies cannot accommodate intents absent from training data, limiting system adaptability to evolving user needs.
Open-world intent detection addresses this limitation by recognizing previously unseen intents through adaptive decision boundaries~\cite{zhang2021deep} and language model representations~\cite{zhang-etal-2021-textoir}.
As real-world interactions grow increasingly complex, users often express multiple interconnected purposes within single utterances rather than isolated intents.
Multi-intent detection emerges to handle this complexity, with benchmarks like MixATIS, MixSNIPS~\cite{qin-etal-2020-agif}, and BlendX~\cite{yoon-etal-2024-blendx} evaluating systems on identifying coexisting intents.
Despite these advances toward handling complex utterances, existing benchmarks evaluate intent identification without user-specific personalization—they assume identical intent interpretations regardless of individual user contexts and histories.
\bench introduces intent identification in \pqa task, where user backgrounds and interaction histories fundamentally shape which intents users prioritize when seeking information.

\subsection{Personalized Question Answering Benchmarks}
\label{subsec:pqa}

\pqa benchmarks evaluate whether systems adapt responses to individual user characteristics or preferences.
Early approaches focus on leveraging user interaction histories for retrieval and ranking.
SE-PQA~\cite{10.1145/3589335.3651445} situates personalization within cQA, leveraging asker–answerer histories to personalize retrieval and ranking.
UQABench~\cite{10.1145/3711896.3737385} assesses compact user embeddings for steering language models toward preference-aligned generation.
PerLTQA~\cite{du-etal-2024-perltqa} measures systems' ability to retrieve and synthesize information from long-term personal memory representations.
Recently, the focus has shifted toward fine-grained content evaluation based on detailed user-provided information.
LaMP-QA~\cite{salemi2025lampqabenchmarkpersonalizedlongform} measures how well generated responses cover aspects extracted from user narratives—detailed descriptions users provide explaining their questions.
However, these benchmarks evaluate retrieval performance or response coverage without measuring intent identification.
Intent identification is critical in \pqa because understanding which motivations drive user questions fundamentally determines whether generated responses satisfy information needs.
Users typically express multiple intents when posing questions, and their answer selection reveals which among these intents they prioritize—intents aligned with selected answers represent what users considered essential for meeting minimum information requirements.
We define these prioritized intents as core intents, and \bench evaluates a system's ability to identify the intents that users  prioritize.

\begin{table*}[t]
\centering
\small
\setlength{\tabcolsep}{2.5pt}
\caption{Answer generation performance with different intent configurations. Each cell represents Intent/Answer performance separated by "/" (*: significant vs. other configurations, paired t-test, $p$ < 0.05).}
\vspace{-5pt}
\begin{tabular}{ll ccc ccc ccc}
    \toprule
    \multirow{2.5}{*}{\textbf{Model}} & \multirow{2.5}{*}{\textbf{Method}} & \multicolumn{3}{c}{\textbf{Entertainment}} & \multicolumn{3}{c}{\textbf{Living}} & \multicolumn{3}{c}{\textbf{Social}} \\
    \cmidrule(lr){3-5} \cmidrule(lr){6-8} \cmidrule(lr){9-11}
    & & Precision & Recall & F1 & Precision & Recall & F1 & Precision & Recall & F1 \\
    \midrule
    \multirow{3.3}{*}{Qwen2.5-7B-It}
    & Predicted 
    & 0.646/0.333 & 0.487/0.269 & 0.528/0.285 & 0.632/0.337 & 0.472/0.310 & 0.517/0.309 & 0.556/0.353 & 0.448/0.309 & 0.469/0.314 \\
    & Noisy  
    & 0.500/\underline{0.462} & 1.000/\underline{0.387} & 0.667/\underline{0.405} & 0.500/\underline{0.450} & 1.000/\underline{0.416} & 0.667/\underline{0.414} & 0.500/\underline{0.513} & 1.000/\underline{0.429} & 0.667/\underline{0.447} \\
    \cmidrule(lr){2-11}
    & Core Only 
    & 1.000/\textbf{0.537*} & 1.000/\textbf{0.404*} & 1.000/\textbf{0.442*} & 1.000/\textbf{0.539*} & 1.000/\textbf{0.449*} & 1.000/\textbf{0.466*} & 1.000/\textbf{0.588*} & 1.000/\textbf{0.457*} & 1.000/\textbf{0.493*} \\
    \midrule[\heavyrulewidth]
    \multirow{3.3}{*}{Gemma3-12B-It}
    & Predicted 
    & 0.613/0.372 & 0.488/0.311 & 0.518/0.322 & 0.599/0.395 & 0.478/0.363 & 0.507/0.361 & 0.551/0.390 & 0.467/0.364 & 0.481/0.358 \\
    & Noisy 
    & 0.500/\underline{0.489} & 1.000/\underline{0.420} & 0.667/\underline{0.434} & 0.500/\underline{0.504} & 1.000/\underline{0.455} & 0.667/\underline{0.457} & 0.500/\underline{0.525} & 1.000/\underline{0.460} & 0.667/\underline{0.471} \\
    \cmidrule(lr){2-11}
    & Core Only 
    & 1.000/\textbf{0.532*} & 1.000/\textbf{0.433*} & 1.000/\textbf{0.459*} & 1.000/\textbf{0.543*} & 1.000/\textbf{0.476*} & 1.000/\textbf{0.486*} & 1.000/\textbf{0.565*} & 1.000/\textbf{0.474} & 1.000/\textbf{0.493*} \\
    \midrule[\heavyrulewidth]
    \multirow{3.3}{*}{GPT-4o-Mini}
    & Predicted 
    & 0.605/0.389 & 0.498/0.300 & 0.517/0.326 & 0.624/0.417 & 0.506/0.357 & 0.530/0.369 & 0.554/0.431 & 0.488/0.343 & 0.493/0.364 \\
    & Noisy 
    & 0.500/\underline{0.528} & 1.000/\underline{0.387} & 0.667/\underline{0.430} & 0.500/\underline{0.525} & 1.000/\underline{0.441} & 0.667/\underline{0.461} & 0.500/\underline{0.590} & 1.000/\underline{0.435} & 0.667/\underline{0.481} \\
    \cmidrule(lr){2-11}
    & Core Only 
    & 1.000/\textbf{0.572*} & 1.000/\textbf{0.407*} & 1.000/\textbf{0.457*} & 1.000/\textbf{0.566*} & 1.000/\textbf{0.459*} & 1.000/\textbf{0.487*} & 1.000/\textbf{0.617*} & 1.000/\textbf{0.452*} & 1.000/\textbf{0.502*} \\
    \bottomrule
\end{tabular}
\label{tab:intent_answer}
\end{table*}

\begin{table}[t]
\centering
\setlength{\tabcolsep}{3pt}
\small
\caption{Performance of \intrag in generating at least one relevant information piece and overall accuracy, conditioned on core intent identification correctness.}
\vspace{-5pt}
\begin{tabular}{llcccc}
    \toprule
    \textbf{Model} & \textbf{Metric} & \textbf{Ent.} & \textbf{Liv.} & \textbf{Soc.} & \textbf{Avg.} \\
    \midrule
    \multirow{4.3}{*}{Qwen2.5-7B-It} & P($\geq$1|\xmark) & 0.3938 & 0.4245 & 0.4477 & 0.4220 \\
    & P($\geq$1|\cmark) & \textbf{0.5861 } & \textbf{0.6482 } & \textbf{0.6623 } & \textbf{0.6322 } \\
    \cmidrule(lr){2-6}
    & Acc|\xmark & 0.2095 & 0.2343 & 0.2480 & 0.2306 \\
    & Acc|\cmark & \textbf{0.3039} & \textbf{0.3658} & \textbf{0.3540} & \textbf{0.3412} \\
    \midrule[\heavyrulewidth]
    \multirow{4.3}{*}{Gemma3-12B-It} & P($\geq$1|\xmark) & 0.4576 & 0.5070 & 0.5153 & 0.4933 \\
    & P($\geq$1|\cmark) & \textbf{0.6638 } & \textbf{0.6928 } & \textbf{0.7657 } & \textbf{0.7074 } \\
    \cmidrule(lr){2-6}
    & Acc|\xmark & 0.2431 & 0.3022 & 0.2921 & 0.2791 \\
    & Acc|\cmark & \textbf{0.3506 } & \textbf{0.3922 } & \textbf{0.4229 } & \textbf{0.3886 } \\
    \midrule[\heavyrulewidth]
    \multirow{4.3}{*}{GPT-4o-Mini} & P($\geq$1|\xmark) & 0.4477 & 0.4920 & 0.4845 & 0.4747 \\
    & P($\geq$1|\cmark) & \textbf{0.6418 } & \textbf{0.7067 } & \textbf{0.7576 } & \textbf{0.7020 } \\
    \cmidrule(lr){2-6}
    & Acc|\xmark & 0.2398 & 0.2706 & 0.2615 & 0.2573 \\
    & Acc|\cmark & \textbf{0.3364 } & \textbf{0.4017 } & \textbf{0.4013 } & \textbf{0.3798 } \\
    \bottomrule
\end{tabular}
\label{tab:intent_answer_cond}
\end{table}

\section{Limitations}
\label{sec:limitations}

\rev{\textbf{LLM-based construction and evaluation.}
Our benchmark relies on LLMs throughout the pipeline---intent generation, quality filtering, core intent selection, and evaluation. While LLMs fundamentally shape the benchmark structure, this design enables scalable construction across thousands of instances where manual annotation alone could not achieve comparable scale or consistency. Our multi-stage quality control---including the Generate-then-Filter methodology with independent verification and human validation on 400 instances---mitigates errors from failure modes such as abstract intents merging distinct motivations or occasional introduction of motivations not explicitly present in narratives, though instances where generated intents are plausible yet subtly misaligned with user motivations may persist. Cross-model consistency (Section 3.2) demonstrates robustness across model families but does not resolve the circularity of LLM-constructed benchmarks evaluating LLMs. Larger-scale human annotation or non-LLM verification remains an important direction for strengthening validity.}

\noindent
\rev{\textbf{Ground-truth validity.}
Our framework derives core intents from selected answers, grounded in satisficing theory where users choose answers meeting their acceptance thresholds~\cite{simon1955behavioral, agosto2002bounded}, with empirical user studies confirming that answer selections reliably reflect prioritized motivations in information-seeking contexts~\cite{agosto2002bounded, chen2012understanding, prabha2007enough}. However, answer selection in community QA platforms may also be influenced by factors such as writing quality, perceived authority of the answerer, response timing, or community voting patterns. While these factors extend beyond direct content-intent alignment, they can also be viewed as part of the broader satisficing process where users weigh multiple aspects when establishing acceptance thresholds. Our quality control procedures mitigate systematic errors, and we believe that selected answers serve as the best available behavioral proxy for user priorities rather than a perfect reflection of true intent. Conducting user studies where users explicitly state their priorities would provide valuable additional validation, though such studies face practical challenges including retrospective recall bias and scalability constraints.}

\noindent
\rev{\textbf{Scope and generalizability.}
IPQA targets community question answering, where users provide explicit textual interaction histories through past questions and narratives. This represents one important and well-studied setting in PQA, but differs from other personalization scenarios: personalized assistants with private memory, systems leveraging implicit signals such as clicks or browsing behavior, and multi-turn conversational contexts. The concept of core intents and our evaluation methodology may generalize, but the specific construction pipeline---relying on explicit narratives and accepted answers---requires adaptation for settings where user intent signals take different forms.}

\section{Conclusion}
\label{sec:conclusion}

This work introduces the concept of core intents in personalized question answering, where users prioritize specific intents when selecting answers to satisfy their information needs.
To evaluate core intent identification, we propose \bench, a benchmark with rigorous quality control and an evaluation framework validated against human judgment.
Experimental results reveal that current language models struggle with core intent identification, failing to capture intent patterns from user histories and showing degraded performance as questions express multiple intents.
While not all intents aligned with selected answers may represent true priorities, this approach provides a more practical framework grounded in observable behavior compared to treating all intents equally without empirical evidence.

\rev{These findings open several research directions for improving core intent identification in future systems. Beyond community QA, the concept of core intents may extend to broader personalization scenarios including recommendation systems and conversational AI, where understanding which motivations users prioritize is fundamental to user satisfaction. Our experiments on intent-answer correlation (Section 3.6) demonstrate that accurate core intent identification directly improves answer generation quality, suggesting a concrete application: incorporating core intent identification as an intermediate reasoning step in PQA pipelines to guide response generation toward user-prioritized information needs. Future work includes validating core intent annotations through user studies with explicitly stated priorities, extending the benchmark to diverse PQA settings with different user signal types, and developing models that jointly identify and leverage core intents for personalized response generation.}

\begin{acks}
This work was supported by the IITP grants funded by the Korea government (MSIT) (No.RS-2020-II201361; RS-2024-00457882, AI Research Hub Project; RS-2026-25520654).
\end{acks}

\bibliographystyle{ACM-Reference-Format}
\bibliography{base}

@inproceedings{zhang-etal-2021-textoir,
    title = "{TEXTOIR}: An Integrated and Visualized Platform for Text Open Intent Recognition",
    author = "Zhang, Hanlei  and
      Li, Xiaoteng  and
      Xu, Hua  and
      Zhang, Panpan  and
      Zhao, Kang  and
      Gao, Kai",
    editor = "Ji, Heng  and
      Park, Jong C.  and
      Xia, Rui",
    booktitle = "Proceedings of the 59th Annual Meeting of the Association for Computational Linguistics and the 11th International Joint Conference on Natural Language Processing: System Demonstrations",
    month = aug,
    year = "2021",
    address = "Online",
    publisher = "Association for Computational Linguistics",
    url = "https://aclanthology.org/2021.acl-demo.20/",
    doi = "10.18653/v1/2021.acl-demo.20",
    pages = "167--174"
}

@article{10.1109/TASLP.2023.3265203,
author = {Zhang, Hanlei and Xu, Hua and Zhao, Shaojie and Zhou, Qianrui},
title = {Learning Discriminative Representations and Decision Boundaries for Open Intent Detection},
year = {2023},
issue_date = {2023},
publisher = {IEEE Press},
volume = {31},
issn = {2329-9290},
url = {https://doi.org/10.1109/TASLP.2023.3265203},
doi = {10.1109/TASLP.2023.3265203},
journal = {IEEE/ACM Trans. Audio, Speech and Lang. Proc.},
month = apr,
pages = {1611–1623},
numpages = {13}
}

@inproceedings{zhang2021deep,
  title={Deep open intent classification with adaptive decision boundary},
  author={Zhang, Hanlei and Xu, Hua and Lin, Ting-En},
  booktitle={Proceedings of the AAAI Conference on Artificial Intelligence},
  volume={35},
  number={16},
  pages={14374--14382},
  year={2021}
}

@inproceedings{casanueva-etal-2020-efficient,
    title = "Efficient Intent Detection with Dual Sentence Encoders",
    author = "Casanueva, I{\~n}igo  and
      Tem{\v{c}}inas, Tadas  and
      Gerz, Daniela  and
      Henderson, Matthew  and
      Vuli{\'c}, Ivan",
    editor = "Wen, Tsung-Hsien  and
      Celikyilmaz, Asli  and
      Yu, Zhou  and
      Papangelis, Alexandros  and
      Eric, Mihail  and
      Kumar, Anuj  and
      Casanueva, I{\~n}igo  and
      Shah, Rushin",
    booktitle = "Proceedings of the 2nd Workshop on Natural Language Processing for Conversational AI",
    month = jul,
    year = "2020",
    address = "Online",
    publisher = "Association for Computational Linguistics",
    url = "https://aclanthology.org/2020.nlp4convai-1.5/",
    doi = "10.18653/v1/2020.nlp4convai-1.5",
    pages = "38--45"
}

@inproceedings{larson-etal-2019-evaluation,
    title = "An Evaluation Dataset for Intent Classification and Out-of-Scope Prediction",
    author = "Larson, Stefan  and
      Mahendran, Anish  and
      Peper, Joseph J.  and
      Clarke, Christopher  and
      Lee, Andrew  and
      Hill, Parker  and
      Kummerfeld, Jonathan K.  and
      Leach, Kevin  and
      Laurenzano, Michael A.  and
      Tang, Lingjia  and
      Mars, Jason",
    editor = "Inui, Kentaro  and
      Jiang, Jing  and
      Ng, Vincent  and
      Wan, Xiaojun",
    booktitle = "Proceedings of the 2019 Conference on Empirical Methods in Natural Language Processing and the 9th International Joint Conference on Natural Language Processing (EMNLP-IJCNLP)",
    month = nov,
    year = "2019",
    address = "Hong Kong, China",
    publisher = "Association for Computational Linguistics",
    url = "https://aclanthology.org/D19-1131/",
    doi = "10.18653/v1/D19-1131",
    pages = "1311--1316"
}

@article{coucke2018snips,
  title={Snips voice platform: an embedded spoken language understanding system for private-by-design voice interfaces},
  author={Coucke, Alice and Saade, Alaa and Ball, Adrien and Bluche, Th{\'e}odore and Caulier, Alexandre and Leroy, David and Doumouro, Cl{\'e}ment and Gisselbrecht, Thibault and Caltagirone, Francesco and Lavril, Thibaut and others},
  journal={arXiv preprint arXiv:1805.10190},
  year={2018}
}

@misc{kumar2024longlampbenchmarkpersonalizedlongform,
      title={LongLaMP: A Benchmark for Personalized Long-form Text Generation}, 
      author={Ishita Kumar and Snigdha Viswanathan and Sushrita Yerra and Alireza Salemi and Ryan A. Rossi and Franck Dernoncourt and Hanieh Deilamsalehy and Xiang Chen and Ruiyi Zhang and Shubham Agarwal and Nedim Lipka and Chien Van Nguyen and Thien Huu Nguyen and Hamed Zamani},
      year={2024},
      eprint={2407.11016},
      archivePrefix={arXiv},
      primaryClass={cs.CL},
      url={https://arxiv.org/abs/2407.11016}, 
}

@inproceedings{salemi2025lampqabenchmarkpersonalizedlongform,
  title={{LaMP-QA}: A Benchmark for Personalized Long-form Question Answering},
  author={Salemi, Alireza and Zamani, Hamed},
  booktitle={Proceedings of the 2025 Conference on Empirical Methods in Natural Language Processing},
  pages={1139--1159},
  year={2025}
}

@inproceedings{post-2018-call,
    title = "A Call for Clarity in Reporting {BLEU} Scores",
    author = "Post, Matt",
    editor = "Bojar, Ond{\v{r}}ej  and
      Chatterjee, Rajen  and
      Federmann, Christian  and
      Fishel, Mark  and
      Graham, Yvette  and
      Haddow, Barry  and
      Huck, Matthias  and
      Yepes, Antonio Jimeno  and
      Koehn, Philipp  and
      Monz, Christof  and
      Negri, Matteo  and
      N{\'e}v{\'e}ol, Aur{\'e}lie  and
      Neves, Mariana  and
      Post, Matt  and
      Specia, Lucia  and
      Turchi, Marco  and
      Verspoor, Karin",
    booktitle = "Proceedings of the Third Conference on Machine Translation: Research Papers",
    month = oct,
    year = "2018",
    address = "Brussels, Belgium",
    publisher = "Association for Computational Linguistics",
    url = "https://aclanthology.org/W18-6319/",
    doi = "10.18653/v1/W18-6319",
    pages = "186--191"
}

@inproceedings{lin-hovy-2003-automatic,
    title = "Automatic Evaluation of Summaries Using N-gram Co-occurrence Statistics",
    author = "Lin, Chin-Yew  and
      Hovy, Eduard",
    booktitle = "Proceedings of the 2003 Human Language Technology Conference of the North {A}merican Chapter of the Association for Computational Linguistics",
    year = "2003",
    url = "https://aclanthology.org/N03-1020/",
    pages = "150--157"
}

@inproceedings{banerjee-lavie-2005-meteor,
    title = "{METEOR}: An Automatic Metric for {MT} Evaluation with Improved Correlation with Human Judgments",
    author = "Banerjee, Satanjeev  and
      Lavie, Alon",
    editor = "Goldstein, Jade  and
      Lavie, Alon  and
      Lin, Chin-Yew  and
      Voss, Clare",
    booktitle = "Proceedings of the {ACL} Workshop on Intrinsic and Extrinsic Evaluation Measures for Machine Translation and/or Summarization",
    month = jun,
    year = "2005",
    address = "Ann Arbor, Michigan",
    publisher = "Association for Computational Linguistics",
    url = "https://aclanthology.org/W05-0909/",
    pages = "65--72"
}

@misc{zhang2020bertscoreevaluatingtextgeneration,
      title={BERTScore: Evaluating Text Generation with BERT}, 
      author={Tianyi Zhang and Varsha Kishore and Felix Wu and Kilian Q. Weinberger and Yoav Artzi},
      year={2020},
      eprint={1904.09675},
      archivePrefix={arXiv},
      primaryClass={cs.CL},
      url={https://arxiv.org/abs/1904.09675}, 
}

@inproceedings{liu-etal-2023-g,
    title = "{G}-Eval: {NLG} Evaluation using Gpt-4 with Better Human Alignment",
    author = "Liu, Yang  and
      Iter, Dan  and
      Xu, Yichong  and
      Wang, Shuohang  and
      Xu, Ruochen  and
      Zhu, Chenguang",
    editor = "Bouamor, Houda  and
      Pino, Juan  and
      Bali, Kalika",
    booktitle = "Proceedings of the 2023 Conference on Empirical Methods in Natural Language Processing",
    month = dec,
    year = "2023",
    address = "Singapore",
    publisher = "Association for Computational Linguistics",
    url = "https://aclanthology.org/2023.emnlp-main.153/",
    doi = "10.18653/v1/2023.emnlp-main.153",
    pages = "2511--2522"
}

@inproceedings{seo-etal-2025-mt,
    title = "{MT}-{RAIG}: Novel Benchmark and Evaluation Framework for Retrieval-Augmented Insight Generation over Multiple Tables",
    author = "Seo, Kwangwook  and
      Kwon, Donguk  and
      Lee, Dongha",
    editor = "Che, Wanxiang  and
      Nabende, Joyce  and
      Shutova, Ekaterina  and
      Pilehvar, Mohammad Taher",
    booktitle = "Proceedings of the 63rd Annual Meeting of the Association for Computational Linguistics (Volume 1: Long Papers)",
    month = jul,
    year = "2025",
    address = "Vienna, Austria",
    publisher = "Association for Computational Linguistics",
    url = "https://aclanthology.org/2025.acl-long.1128/",
    doi = "10.18653/v1/2025.acl-long.1128",
    pages = "23142--23172",
    ISBN = "979-8-89176-251-0"
}

@inproceedings{10.5555/3737916.3738608,
author = {Ru, Dongyu and Qiu, Lin and Hu, Xiangkun and Zhang, Tianhang and Shi, Peng and Chang, Shuaichen and Jiayang, Cheng and Wang, Cunxiang and Sun, Shichao and Li, Huanyu and Zhang, Zizhao and Wang, Binjie and Jiang, Jiarong and He, Tong and Wang, Zhiguo and Liu, Pengfei and Zhang, Yue and Zhang, Zheng},
title = {RAGCHECKER: a fine-grained framework for diagnosing retrieval-augmented generation},
year = {2025},
isbn = {9798331314385},
publisher = {Curran Associates Inc.},
address = {Red Hook, NY, USA},
booktitle = {Proceedings of the 38th International Conference on Neural Information Processing Systems},
articleno = {692},
numpages = {29},
location = {Vancouver, BC, Canada},
series = {NIPS '24}
}

@misc{heo2025largelanguagemodelseffective,
      title={Can Large Language Models be Effective Online Opinion Miners?}, 
      author={Ryang Heo and Yongsik Seo and Junseong Lee and Dongha Lee},
      year={2025},
      eprint={2505.15695},
      archivePrefix={arXiv},
      primaryClass={cs.CL},
      url={https://arxiv.org/abs/2505.15695}, 
}

@String{Computing = "Computing" }

@ArtifactSoftware{R,
    title = {R: A Language and Environment for Statistical Computing},
    author = {{R Core Team}},
    organization = {R Foundation for Statistical Computing},
    address = {Vienna, Austria},
    year = {2019},
    url = {https://www.R-project.org/},
}

@article{salemi2023lamp,
  title={Lamp: When large language models meet personalization},
  author={Salemi, Alireza and Mysore, Sheshera and Bendersky, Michael and Zamani, Hamed},
  journal={arXiv preprint arXiv:2304.11406},
  year={2023}
}

@inproceedings{10.1145/3589335.3651445,
author = {Kasela, Pranav and Braga, Marco and Pasi, Gabriella and Perego, Raffaele},
title = {SE-PQA: Personalized Community Question Answering},
year = {2024},
isbn = {9798400701726},
publisher = {Association for Computing Machinery},
address = {New York, NY, USA},
url = {https://doi.org/10.1145/3589335.3651445},
doi = {10.1145/3589335.3651445},
booktitle = {Companion Proceedings of the ACM Web Conference 2024},
pages = {1095–1098},
numpages = {4},
location = {Singapore, Singapore},
series = {WWW '24}
}

@inproceedings{du-etal-2024-perltqa,
    title = "{P}er{LTQA}: A Personal Long-Term Memory Dataset for Memory Classification, Retrieval, and Fusion in Question Answering",
    author = "Du, Yiming  and
      Wang, Hongru  and
      Zhao, Zhengyi  and
      Liang, Bin  and
      Wang, Baojun  and
      Zhong, Wanjun  and
      Wang, Zezhong  and
      Wong, Kam-Fai",
    editor = "Wong, Kam-Fai  and
      Zhang, Min  and
      Xu, Ruifeng  and
      Li, Jing  and
      Wei, Zhongyu  and
      Gui, Lin  and
      Liang, Bin  and
      Zhao, Runcong",
    booktitle = "Proceedings of the 10th SIGHAN Workshop on Chinese Language Processing (SIGHAN-10)",
    month = aug,
    year = "2024",
    address = "Bangkok, Thailand",
    publisher = "Association for Computational Linguistics",
    url = "https://aclanthology.org/2024.sighan-1.18/",
    pages = "152--164"
}

@inproceedings{10.1145/3711896.3737385,
author = {Liu, Langming and Liu, Shilei and Yuan, Yujin and Zhang, Yizhen and Yan, Bencheng and Zeng, Zhiyuan and Wang, Zihao and Liu, Jiaqi and Wang, Di and Su, Wenbo and Wang, Pengjie and Xu, Jian and Zheng, Bo},
title = {UQABench: Evaluating User Embedding for Prompting LLMs in Personalized Question Answering},
year = {2025},
isbn = {9798400714542},
publisher = {Association for Computing Machinery},
address = {New York, NY, USA},
url = {https://doi.org/10.1145/3711896.3737385},
doi = {10.1145/3711896.3737385},
booktitle = {Proceedings of the 31st ACM SIGKDD Conference on Knowledge Discovery and Data Mining V.2},
pages = {5652–5661},
numpages = {10},
location = {Toronto ON, Canada},
series = {KDD '25}
}

@article{hui2024qwen2,
  title={Qwen2. 5-coder technical report},
  author={Hui, Binyuan and Yang, Jian and Cui, Zeyu and Yang, Jiaxi and Liu, Dayiheng and Zhang, Lei and Liu, Tianyu and Zhang, Jiajun and Yu, Bowen and Lu, Keming and others},
  journal={arXiv preprint arXiv:2409.12186},
  year={2024}
}

@article{dubey2024llama,
  title={The llama 3 herd of models},
  author={Dubey, Abhimanyu and Jauhri, Abhinav and Pandey, Abhinav and Kadian, Abhishek and Al-Dahle, Ahmad and Letman, Aiesha and Mathur, Akhil and Schelten, Alan and Yang, Amy and Fan, Angela and others},
  journal={arXiv e-prints},
  pages={arXiv--2407},
  year={2024}
}

@article{team2025gemma,
  title={Gemma 3 technical report},
  author={Team, Gemma and Kamath, Aishwarya and Ferret, Johan and Pathak, Shreya and Vieillard, Nino and Merhej, Ramona and Perrin, Sarah and Matejovicova, Tatiana and Ram{\'e}, Alexandre and Rivi{\`e}re, Morgane and others},
  journal={arXiv preprint arXiv:2503.19786},
  year={2025}
}

@article{zhang2025qwen3,
  title={Qwen3 Embedding: Advancing Text Embedding and Reranking Through Foundation Models},
  author={Zhang, Yanzhao and Li, Mingxin and Long, Dingkun and Zhang, Xin and Lin, Huan and Yang, Baosong and Xie, Pengjun and Yang, An and Liu, Dayiheng and Lin, Junyang and others},
  journal={arXiv preprint arXiv:2506.05176},
  year={2025}
}

@inproceedings{qin-etal-2020-agif,
    title = "{AGIF}: An Adaptive Graph-Interactive Framework for Joint Multiple Intent Detection and Slot Filling",
    author = "Qin, Libo  and
      Xu, Xiao  and
      Che, Wanxiang  and
      Liu, Ting",
    editor = "Cohn, Trevor  and
      He, Yulan  and
      Liu, Yang",
    booktitle = "Findings of the Association for Computational Linguistics: EMNLP 2020",
    month = nov,
    year = "2020",
    address = "Online",
    publisher = "Association for Computational Linguistics",
    url = "https://aclanthology.org/2020.findings-emnlp.163/",
    doi = "10.18653/v1/2020.findings-emnlp.163",
    pages = "1807--1816"
}

@inproceedings{yoon-etal-2024-blendx,
    title = "{B}lend{X}: Complex Multi-Intent Detection with Blended Patterns",
    author = "Yoon, Yejin  and
      Lee, Jungyeon  and
      Kim, Kangsan  and
      Park, Chanhee  and
      Kim, Taeuk",
    editor = "Calzolari, Nicoletta  and
      Kan, Min-Yen  and
      Hoste, Veronique  and
      Lenci, Alessandro  and
      Sakti, Sakriani  and
      Xue, Nianwen",
    booktitle = "Proceedings of the 2024 Joint International Conference on Computational Linguistics, Language Resources and Evaluation (LREC-COLING 2024)",
    month = may,
    year = "2024",
    address = "Torino, Italia",
    publisher = "ELRA and ICCL",
    url = "https://aclanthology.org/2024.lrec-main.218/",
    pages = "2428--2439"
}

@inproceedings{chen2012understanding,
  title={Understanding user intent in community question answering},
  author={Chen, Long and Zhang, Dell and Mark, Levene},
  booktitle={Proceedings of the 21st international conference on world wide web},
  pages={823--828},
  year={2012}
}

@article{prabha2007enough,
  title={What is enough? Satisficing information needs},
  author={Prabha, Chandra and Silipigni Connaway, Lynn and Olszewski, Lawrence and Jenkins, Lillie R},
  journal={Journal of documentation},
  volume={63},
  number={1},
  pages={74--89},
  year={2007},
  publisher={Emerald Group Publishing Limited}
}

@article{agosto2002bounded,
  title={Bounded rationality and satisficing in young people's Web-based decision making},
  author={Agosto, Denise E},
  journal={Journal of the American society for Information Science and Technology},
  volume={53},
  number={1},
  pages={16--27},
  year={2002},
  publisher={Wiley Online Library}
}

@article{simon1955behavioral,
  title={A behavioral model of rational choice},
  author={Simon, Herbert A},
  journal={The quarterly journal of economics},
  pages={99--118},
  year={1955},
  publisher={JSTOR}
}

@inproceedings{quarteroni-2010-personalized,
    title = "Personalized Question Answering",
    author = "Quarteroni, Silvia",
    editor = "Daille, B{\'e}atrice  and
      Villemonte de la Clergerie, {\'E}ric  and
      Lepage, Yves  and
      Yvon, Fran{\c{c}}ois",
    booktitle = "Traitement Automatique des Langues, Volume 51, Num{\'e}ro 1 : Varia [Varia]",
    year = "2010",
    address = "France",
    publisher = "ATALA (Association pour le Traitement Automatique des Langues)",
    url = "https://aclanthology.org/2010.tal-1.4/",
    pages = "97--123"
}

@article{klie-etal-2024-analyzing,
    title = "Analyzing Dataset Annotation Quality Management in the Wild",
    author = "Klie, Jan-Christoph  and
      Eckart de Castilho, Richard  and
      Gurevych, Iryna",
    journal = "Computational Linguistics",
    volume = "50",
    number = "3",
    month = sep,
    year = "2024",
    address = "Cambridge, MA",
    publisher = "MIT Press",
    url = "https://aclanthology.org/2024.cl-3.1/",
    doi = "10.1162/coli_a_00516",
    pages = "817--866"
}

@inproceedings{salemi-etal-2025-expert,
    title = "{E}x{P}er{T}: Effective and Explainable Evaluation of Personalized Long-Form Text Generation",
    author = "Salemi, Alireza  and
      Killingback, Julian  and
      Zamani, Hamed",
    editor = "Che, Wanxiang  and
      Nabende, Joyce  and
      Shutova, Ekaterina  and
      Pilehvar, Mohammad Taher",
    booktitle = "Findings of the Association for Computational Linguistics: ACL 2025",
    month = jul,
    year = "2025",
    address = "Vienna, Austria",
    publisher = "Association for Computational Linguistics",
    url = "https://aclanthology.org/2025.findings-acl.900/",
    doi = "10.18653/v1/2025.findings-acl.900",
    pages = "17516--17532",
    ISBN = "979-8-89176-256-5"
}

@inproceedings{kwon2023efficient,
  title={Efficient memory management for large language model serving with pagedattention},
  author={Kwon, Woosuk and Li, Zhuohan and Zhuang, Siyuan and Sheng, Ying and Zheng, Lianmin and Yu, Cody Hao and Gonzalez, Joseph and Zhang, Hao and Stoica, Ion},
  booktitle={Proceedings of the 29th symposium on operating systems principles},
  pages={611--626},
  year={2023}
}

@misc{kim2025rpmreasoninglevelpersonalizationblackbox,
      title={RPM: Reasoning-Level Personalization for Black-Box Large Language Models}, 
      author={Jieyong Kim and Tongyoung Kim and Soojin Yoon and Jaehyung Kim and Dongha Lee},
      year={2025},
      eprint={2505.21082},
      archivePrefix={arXiv},
      primaryClass={cs.CL},
      url={https://arxiv.org/abs/2505.21082}, 
}

@inproceedings{10.1145/3726302.3730055,
author = {Kim, Jieyong and Kim, Hyunseo and Cho, Hyunjin and Kang, SeongKu and Chang, Buru and Yeo, Jinyoung and Lee, Dongha},
title = {Review-driven Personalized Preference Reasoning with Large Language Models for Recommendation},
year = {2025},
isbn = {9798400715921},
publisher = {Association for Computing Machinery},
address = {New York, NY, USA},
url = {https://doi.org/10.1145/3726302.3730055},
doi = {10.1145/3726302.3730055},
booktitle = {Proceedings of the 48th International ACM SIGIR Conference on Research and Development in Information Retrieval},
pages = {1697–1706},
numpages = {10},
location = {Padua, Italy},
series = {SIGIR '25}
}

\appendix

\section{Human Evaluation Details}
\label{apdx:annotation}

Following the setup described in Section~\ref{subsec:manual_validation}, annotators first completed structured training on criterion definitions with worked examples. Calibration was conducted on 10\% of instances per task, achieving over 80\% inter-annotator agreement before the main annotation. For dataset validation, annotators received the question, narrative, selected answer, generated intents, and extracted information pieces without indication of LLM generation source, ensuring criterion-based assessment free from confirmation bias. For meta-evaluation (Section~\ref{subsec:meta_evaluation} and Section~\ref{apdx:metaeval_answer}), annotators performed pairwise comparison on completeness and faithfulness dimensions.

\section{Benchmark Domain Distribution}
\label{apdx:benchmark_details}

The dataset spans 47 fine-grained categories across three domain groups.
\textbf{Entertainment} (10 categories, 3,005 instances) is led by gaming (1,231; 41.0\%), rpg (745; 24.8\%), scifi (377; 12.6\%), movies (186; 6.2\%), and boardgames (169; 5.6\%), with remaining categories including music, literature, anime, sound, and musicfans.
\textbf{Living} (18 categories, 2,228 instances) is led by travel (535; 24.0\%), diy (411; 18.5\%), workplace (239; 10.7\%), gardening (155; 7.0\%), and cooking (152; 6.8\%), covering practical domains such as parenting, bicycles, pets, fitness, and health.
\textbf{Social} (19 categories, 2,497 instances) is led by judaism (370; 14.8\%), money (337; 13.5\%), academia (328; 13.1\%), christianity (208; 8.3\%), and law (181; 7.3\%), spanning religious, professional, and interpersonal topics.
Complete category-level statistics are available in our repository.

\section{Answer Quality Meta-Evaluation}
\label{apdx:metaeval_answer}

While IPQA-Eval validates intent identification quality, answer generation quality requires separate validation since the evaluation target differs.
IPQA-Eval (Answer) follows the same protocol as Section~\ref{subsec:meta_evaluation}, applied to answer pairs instead of intent pairs.
Annotators completed the same training procedure described in Appendix~\ref{apdx:annotation} before conducting pairwise comparisons.
Table~\ref{tbl:metaeval_answer} confirms strong human correlation, outperforming all baseline metrics.
These results validate that information piece-based evaluation reliably assesses answer generation quality in PQA contexts.

\begin{table}[h]
\centering
\small
\caption{Qualitative example of intent-answer correlation.}
\label{tab:qualitative}
\vspace{-5pt}
\begin{tabular}{p{0.92\linewidth}}
\toprule
\textbf{Question}: How to introduce ``unnatural'' magic in a high-magic RPG? \\
\midrule
\textbf{Selected answer}: Contrast with normal magic; rule-breaking; player-driven discovery \\
\midrule
\textbf{Predicted intent}: ``Balanced method for incorporating magical elements'' \\
\textbf{$\rightarrow$ Generated answer}: Generic design considerations (Source and origin, narrative integration, balance) \\
\midrule
\textbf{Ground truth intents}: Contrast; coherence; player discovery \\
\textbf{$\rightarrow$ Generated answer}: Targeted mechanics (Rule-breaking mechanics, perceptible differences, discovery techniques) \\
\bottomrule
\end{tabular}
\end{table}

\begin{table}[h]
\small
\centering
\caption{Meta-Evaluation results for information pieces.}
\label{tbl:metaeval_answer}
\vspace{-5pt}
\begin{tabular}{lcc}
    \toprule
    \textbf{Evaluation Metric} & \textbf{Completeness} & \textbf{Faithfulness} \\ 
    \midrule[\heavyrulewidth]
    SacreBLEU~\cite{post-2018-call}                                     & 34.86        & 23.16        \\
    ROUGE-L~\cite{lin-hovy-2003-automatic}                                    & 36.68        & 20.03        \\
    METEOR~\cite{banerjee-lavie-2005-meteor}                                          & 35.18        & 31.11        \\
    BERTScore~\cite{zhang2020bertscoreevaluatingtextgeneration}                                   & 36.97        & 34.96        \\
    G-Eval~\cite{liu-etal-2023-g}                                      & \underline{61.56}        & \underline{52.28}        \\
    \midrule
    \eval (Answer)                             & \textbf{67.03}        & \textbf{67.82}        \\
    \midrule
    Inter-Human Correlation                     & 82.36        & 79.16        \\ 
    \bottomrule
\end{tabular}
\end{table}

\begin{table}[h]
\centering
\caption{Example instance from \bench.}
\vspace{-5pt}
\label{tbl:case_study}
\footnotesize
\begin{tabular}{p{0.93\columnwidth}}
\toprule
\textbf{Question:} ``How late is too late to describe a character?'' \\
\midrule
\textbf{Narrative (excerpt):} ``...I never list physical traits directly. Later, another character mentions she has red hair. Will this disrupt the mental image readers already formed? Should I mention it earlier?'' \\
\midrule
\textbf{Selected Answer (excerpt):} ``Describe the red hair early and explicitly. Don't feel obligated to add all other physical traits; if red hair is the only relevant trait, make it the only one you describe...'' \\
\midrule
\textbf{Core Intents:} (1) \textit{Preserve reader mental image/continuity}; (2) \textit{Determine optimal timing for revealing details} \\
\midrule
\textbf{Information Pieces} $\rightarrow$ \textbf{Intent:} \\
\hspace{0.5em}$\bullet$ ``Describe red hair early and explicitly'' $\rightarrow$ (2) \\
\hspace{0.5em}$\bullet$ ``Limit physical details to relevant traits'' $\rightarrow$ (2) \\
\hspace{0.5em}$\bullet$ ``Readers remember emphasized non-physical traits'' $\rightarrow$ (1) \\
\hspace{0.5em}$\bullet$ ``Buried physical details get missed'' $\rightarrow$ (1) \\
\hspace{0.5em}$\bullet$ ``Plot relevance determines whether discrepancies matter'' $\rightarrow$ (1) \\
\bottomrule
\end{tabular}
\end{table}

\section{Case Study}
\label{apdx:case_study}

Table~\ref{tbl:case_study} illustrates the annotation structure through a representative instance from the writers category. The question initially yields multiple intents, of which two survive alignment assessment: (1) preserving reader mental image and (2) determining optimal timing for revealing details. The selected answer decomposes into five information pieces, with intent (1) aligning with three pieces on reader perception and intent (2) with two on description strategy. For brevity, the table shows only core intent names; complete annotations are available in our dataset.

\end{document}